\pdfoutput=1
\documentclass{article}

\usepackage[preprint]{neurips_2026}

\usepackage[utf8]{inputenc}
\usepackage[T1]{fontenc}
\usepackage{hyperref}
\hypersetup{hidelinks}
\usepackage{url}
\usepackage{booktabs}
\usepackage{amsmath,amssymb,amsfonts}
\usepackage{nicefrac}
\usepackage{microtype}
\usepackage{xcolor}
\usepackage{graphicx}
\usepackage{multirow}

\newcommand{\R}{\mathbb{R}}
\newcommand{\ours}{\textbf{Ours}}
\title{Measurement-Triggered Channel Memories for\\ Irregularly Sampled Clinical\\ and Wearable Time Series}

\author{%
  Tero Keski-Valkama\\
  Hypersleep (hypersleep.ai)\\
  \texttt{tero.keski-valkama@hypersleep.ai}
}

\begin{document}

\maketitle

\begin{abstract}
Irregularly sampled multivariate time series record each channel at its own times, and whether and when a channel is measured carries information. We classify such records with a small model that processes the record as a sequence of measurement events. Every channel owns a slot that is written only when that channel is measured; it keeps exact running statistics of the channel, a decaying value vector and the time of its last measurement. Each measurement is compared with learned thresholds in its own channel's units before any cross-channel mixing, and a shared temporal state of decaying and rotating modes is carried over the actual elapsed time between measurements. On the five standard splits of the P19 sepsis benchmark, the model reaches a test AUPRC of $63.9 \pm 3.9$, the highest published under this protocol (next: $62.3$, with a pretrained image encoder), and an AUROC of $91.6 \pm 2.2$, equal to the best published value under this protocol, with 62,681 parameters trained on one CPU thread. Paired over the same splits, removing the temporal state and the decaying value vectors from the readout costs AUPRC on every split ($6.7$ points on average), and the model leads gradient-boosted trees fitted to raw per-channel summaries by $2.1 \pm 1.8$ AUPRC points, on four of five splits; an independent rerun on a second host leads them on all five ($+2.6 \pm 1.4$). Those trees score AUROC $91.4$ and AUPRC $61.8$, a baseline the field should report: only two published five-split deep models exceed them, each on one metric and each with a pretrained image encoder. Under the same protocol, among models trained on the benchmark alone, ours is level with the best published PAMAP2 accuracy ($97.8 \pm 0.7$ against $97.7$) and behind the best P12 mortality AUROC ($87.1$ against $88.0$); models with pretrained image encoders report $98.3$ and $88.5$. All results are research benchmark results on public retrospective data.
\end{abstract}

\section{Introduction}

We hypothesise that much of the world is a set of hidden processes running in parallel and linked by causal interactions, observed through asynchronous sequences of events. Sparse, asynchronous and causal models built like those processes should then have a better inductive bias for such data than dense models, and should spend work only where events happen~\citep{keskivalkama2021sleeping}. Irregularly sampled clinical records test the first of these consequences directly.

Intensive-care records, wearable sensors and industrial monitors share a structure: many channels, each measured at its own irregular times, with long silences in between. Whether and when a channel is measured is itself informative, since a laboratory value is often ordered because a clinician expected it to matter, and the time since a channel was last measured says how current its information is.

Recent classifiers for irregularly sampled multivariate time series (IMTS) mostly turn the record into a dense object first: a channel-by-time grid with masks processed by attention or graph mixing~\begin{NoHyper}\citep{zhang2022raindrop,zhang2023warpformer,zhong2025mtm}\end{NoHyper}, tokens or patches for transformers~\begin{NoHyper}\citep{wei2023coformer,zhang2024tpatchgnn}\end{NoHyper}, or rendered line-graph images for vision models~\begin{NoHyper}\citep{li2023vitst,yu2026vire}\end{NoHyper}. The resulting models are accurate, but their cost grows with the grid, and the information carried by missingness and elapsed time has to be relearned from masks and time encodings.

We take the opposite route and keep the record as events. Three components do the work:
\begin{itemize}
    \item \textbf{Measurement-triggered channel slots.} Each channel has a slot written only when that channel is measured. It holds exact running statistics of the channel's values, a non-negative value vector that decays with the time since the measurement that wrote it, and the time of the last measurement, so the readout sees how long each channel has been silent and whether it was ever measured.
    \item \textbf{Channel-typed comparisons.} Each measurement passes learned soft thresholds in its own channel's units before it is mixed with other channels, so channels with different units never share a threshold.
    \item \textbf{An elapsed-time temporal state.} A shared state of decaying and rotating complex modes is carried over the actual gap since the previous measurement and then absorbs the new measurement through a gated update.
\end{itemize}

\paragraph{Findings.}
\begin{itemize}
    \item On P19~\citep{reyna2020sepsis} under the five-split protocol of \citet{zhang2022raindrop}, the model's AUPRC ($63.9 \pm 3.9$) is the highest published and its AUROC ($91.6 \pm 2.2$) equals the best published value (Table~\ref{tab:p19}). Both previous best values come from models with externally pretrained image encoders; among the models of Table~\ref{tab:p19} trained on the benchmark alone, the best reach $59.4$ AUPRC and $90.3$ AUROC. The model has 62,681 parameters and trains in about 12 minutes per split on one CPU thread.
    \item Much of the P19 signal lies in per-channel summary statistics. Gradient-boosted trees on raw per-channel summaries reach $91.4$ / $61.8$; among published five-split results only two deep models exceed them, each on one metric and both with pretrained image encoders. The temporal state and the decaying value vectors still add information the statistics lack: paired over the five splits, they contribute $+6.7$ AUPRC points within the same network, and the model leads the trees by $2.1 \pm 1.8$ AUPRC points on average, on four of five splits (on split 4, $62.92$ against $62.94$). An independent rerun on a second host, which selected epoch 7 instead of 6 on split 4, leads the trees on all five ($+2.6 \pm 1.4$; Table~\ref{tab:compl}, Figure~\ref{fig:paired}).
    \item The same architecture scores $97.8 \pm 0.7$ accuracy on PAM~\citep{reiss2012pamap2} with 46,316 parameters, level with SPECTRA ($97.7 \pm 0.2$), the best published five-split result trained on the benchmark alone, and above MTM~\citep{zhong2025mtm}; two methods with pretrained image encoders report $98.3$. On P12~\citep{silva2012p12} it scores $87.1 \pm 1.1$ AUROC, behind MTM's $88.0$, the best five-split result trained on the benchmark alone, and ViRe's $88.5$ (Section~\ref{sec:pam}).
\end{itemize}

\section{Related work}\label{sec:related}

\paragraph{IMTS classification.} GRU-D~\citep{che2018grud} decays inputs and hidden states toward defaults with the elapsed time, and time-aware~\citep{baytas2017tlstm} and phased~\citep{neil2016phased} recurrent units discount or gate memory by time. Interpolation-prediction networks~\citep{shukla2019ipnet}, set functions~\citep{horn2020seft} and multi-time attention~\citep{shukla2021mtand} avoid a fixed grid. Raindrop~\citep{zhang2022raindrop} introduced the P19/P12/PAM five-split protocol with a graph of sensor dependencies; later work added continuous-time attention~\citep{chen2023contiformer}, compatible transformers~\citep{wei2023coformer}, multi-scale warping~\citep{zhang2023warpformer}, line-graph images classified by vision transformers~\citep{li2023vitst}, graph forecasting models adapted to classification~\citep{yalavarthi2024grafiti,zhang2024tpatchgnn}, channel harmonisation~\citep{liu2025timecheat}, pre-trained language models~\citep{zhang2025istsplm}, joint sequence and image self-supervision~\citep{chen2025seqimage}, multi-scale token mixing~\citep{zhong2025mtm} and vision-informed retrieval~\citep{yu2026vire}. KEDGN~\citep{luo2024kedgn} uses textual variable knowledge in a dynamic graph, Hi-Patch~\citep{luo2025hipatch} hierarchical patch graphs and DBGL~\citep{chen2026dbgl} decay-aware bipartite graphs; KEDGN and Hi-Patch report results on a single split with five seeds, and DBGL in the same setting (Section~\ref{sec:protocol}).

\paragraph{Continuous-time state.} Latent ODEs~\citep{rubanova2019latent}, GRU-ODE-Bayes~\citep{debrouwer2019gruodebayes}, neural CDEs~\citep{kidger2020ncde} and continuous recurrent units~\citep{schirmer2022cru} evolve a hidden state between irregular observations. Diagonal linear state-space layers with complex eigenvalues~\citep{smith2023s5,orvieto2023lru} admit exact transport over arbitrary gaps; our temporal state is of this kind, updated at measurement events rather than on a grid.

\paragraph{Summary statistics.} Hand-built per-channel statistics with boosted trees are a long-standing strong baseline on clinical data; \citet{nie2026statistical} report XGBoost on statistical features above several deep IMTS models under a different, cross-validated protocol. Section~\ref{sec:compl} measures this baseline on the standard splits and asks what a temporal model adds beyond it.

\section{Method}\label{sec:method}

A record is a sequence of time steps $t_1 < \dots < t_L$. At step $n$ a subset $O_n$ of the $C$ channels is measured with values $x_{n,c}$, $c \in O_n$; a record also has static covariates $s$. A channel whose training values are all non-negative is mapped by $\log(1+x)$; every channel is then standardised with the mean and standard deviation of its training measurements and clipped to $[-6, 6]$, giving $z_{n,c}$. Unmeasured channel-steps are never filled in.

\paragraph{Channel-typed comparisons.} A measured value is compared with $J$ learned thresholds of its own channel,
\begin{equation}
\phi_c(z) = \big[z,\ \sigma\big((z-\theta_{c,1})/s_{c,1}\big), \dots, \sigma\big((z-\theta_{c,J})/s_{c,J}\big)\big] \in \R^{J+1},
\end{equation}
with thresholds $\theta_{c,j}$ and positive scales $s_{c,j}$ per channel, and mapped by a channel-specific linear map to features $e_{n,c} = P_c\,\phi_c(z_{n,c}) + b_c \in \R^d$. Only after this step are channels combined: the input of step $n$ is
\begin{equation}
u_n = \sum_{c \in O_n} e_{n,c} + W_g\big[\log(1+\Delta_n),\ |O_n|/C\big], \qquad \Delta_n = t_n - t_{n-1},\ \Delta_1 = 0.
\end{equation}

\paragraph{Temporal state.} A level holds $K$ complex modes $h \in \mathbb{C}^K$ with learned decay rates $r_k > 0$ and angular frequencies $\omega_k$. At step $n$ the state is first carried over the elapsed time and then updated with a gated increment,
\begin{equation}
h_{n,k} = e^{(-r_k + \mathrm{i}\,\omega_k)\Delta_n}\, h_{n-1,k} + g_{n,k}\big(a_{n,k} + \mathrm{i}\, b_{n,k}\big), \qquad g_n = \sigma(G u_n),\quad [a_n; b_n] = A u_n ,
\end{equation}
where $\mathrm{i}$ is the imaginary unit and the same gate multiplies the real and imaginary parts of a mode. The level output is $y_n = \mathrm{LN}\big(u_n + W_o[\mathrm{Re}\,h_n; \mathrm{Im}\,h_n]\big)$ followed by $y_n \leftarrow \mathrm{LN}\big(y_n + \mathrm{MLP}(y_n)\big)$, and it is the input of the next level. Transport over any gap is a multiplication, so the state can be read at any time without inventing an observation. We use two levels of $K = 16$ modes with width $d = 32$; rates are initialised log-spaced between $10^{-2}$ and $10$ per time unit.

\paragraph{Measurement-triggered channel slots.} Slot $c$ is written only at steps with $c \in O_n$. It accumulates the count, mean, minimum, maximum, first and last value of the channel and the last-minus-first change; for PAM also the standard deviation and the mean absolute change between consecutive measurements. It stores the time of the last measurement, from which the readout receives the age $\log(1 + t_L - \tau_c)$, and a presence indicator that separates a never-measured channel from a measured zero. It also holds a value vector of $d_v = 4$ components, overwritten at each measurement with $\mathrm{softplus}(V e_{n,c})$ and read as $v_c \odot \exp(-\kappa\,(t - \tau_c))$ with learned positive rates $\kappa$; a channel measured long ago still contributes through its slow components. The statistics of all channels are normalised jointly and mapped by a linear layer and a GELU.

\paragraph{Readout and training.} A two-layer classifier receives the last-level output at the final step, its mean over the record, a linear map of the static covariates, the value vectors, ages and presence indicators of all channels, and the statistics features. Binary tasks use cross-entropy with the positive class weighted by the training ratio of negatives to positives; PAM uses categorical cross-entropy. For PAM, training records are varied by keeping a random contiguous 80\% of the time steps (times counted from the start of the kept part) and multiplying each channel of a record by a random gain $1 + 0.1\,\epsilon$, $\epsilon \sim \mathcal{N}(0,1)$, and the model scored is an exponential moving average of the weights with factor $0.999$. All hyperparameters are in Appendix~\ref{app:impl}.

\section{Experimental protocol}\label{sec:protocol}

\begin{table}[t]
\centering\small
\caption{Datasets in the Raindrop releases~\citep{zhang2022raindrop}. Observed: fraction of channel-steps with a measurement.}
\label{tab:data}
\begin{tabular}{@{}lrrrrrrl@{}}
\toprule
 & Records & Channels & Static & Max.\ steps & Observed & Positives & Splits (train/val/test) \\
\midrule
P19 & 38,803 & 34 & 6 & 60 & 19.5\% & 4.2\% & 31,042 / 3,880 / 3,881 \\
P12 & 11,988 & 36 & 9 & 214 & 15.1\% & 14.2\% & 9,590 / 1,199 / 1,199 \\
PAM & 5,333 & 17 & -- & 600 & 39.7\% & 8 classes & 4,266 / 533 / 534 \\
\bottomrule
\end{tabular}
\end{table}

\paragraph{Data.} We use the processed releases of \citet{zhang2022raindrop} (Table~\ref{tab:data}). P19 holds the PhysioNet/Computing in Cardiology Challenge 2019 records~\citep{reyna2020sepsis,goldberger2000physionet} with up to 60 hourly steps and the release's binary sepsis label, which \citet{zhang2022raindrop} describe as the occurrence of sepsis within the next 6 hours. P12 holds the 2012 challenge records~\citep{silva2012p12} over the first 48 hours with the in-hospital mortality label.\footnote{\citet{zhang2022raindrop} describe the P12 label as an ICU stay longer than three days (about 93\% positive). The release also holds in-hospital death (14.2\% positive), the label used here and the only one consistent with the P12 AUPRC values reported by the compared methods.} PAM is PAMAP2 activity monitoring~\citep{reiss2012pamap2}: 600-step segments from 17 inertial-sensor channels (temperature, two three-axis accelerometers, a three-axis gyroscope, a three-axis magnetometer and a four-component orientation) in eight activity classes, with 60\% of the values removed at random by the release. The releases store a missing value as zero, so a channel-step counts as measured when its value is non-zero. These are research benchmarks: the P19 release is not the challenge's hidden test set or its utility score.

\paragraph{Protocol.} Each dataset has five fixed splits (80/10/10) defined by the release. For every split we train one model (seed 0) on the training part, select the checkpoint on the validation part (AUROC for P19 and P12, accuracy for PAM) and score the test part once. We report the mean and the sample standard deviation over the five test parts; the references report mean $\pm$ standard deviation over the same splits. The configurations were fixed and recorded before any of their five-split runs: P19 and P12 from development on split-1 validation data, PAM from development on split-1 validation data after its own round of design changes (Appendix~\ref{app:dev}).

\paragraph{References and comparability.} We compare with results reported on the same five splits and metrics: the values compiled by MTM~\citep{zhong2025mtm}, the results of papers that state they follow these splits, ViTST's settings or MTM's protocol~\citep{zhang2025istsplm,chen2025seqimage,yu2026vire}, and those of TimeCHEAT~\citep{liu2025timecheat} (fixed 80/10/10 splits), whose baseline values equal the five-split values, and SPECTRA~\citep{qin2026spectra} (five 8:1:1 folds), most of whose baseline values do. KEDGN~\citep{luo2024kedgn} trains and tests on one of the five splits with five seeds (its public code loads split 1 only) and re-runs its baselines under that setting, where the same baselines score differently from their five-split values (Raindrop: $89.4$ / $61.2$ against $87.0$ / $51.8$); we list their results separately. Hi-Patch~\citep{luo2025hipatch} uses the same setting (its public code also loads split 1 only), and DBGL reports in the same setting, with KEDGN's baseline values. On split 1, the split KEDGN and Hi-Patch use, five seeds of our protocol configuration score $93.4 \pm 0.5$ AUROC and $69.7 \pm 1.8$ AUPRC (seed 0, the protocol model, $94.2$ / $70.3$), against $92.3 \pm 1.0$ / $62.5 \pm 0.7$ for KEDGN (its variable-name variant), $92.1 \pm 0.4$ / $61.1 \pm 2.1$ for Hi-Patch and $93.3 \pm 0.5$ / $66.3 \pm 1.4$ for DBGL. WaveGNN~\citep{hajisafi2025wavegnn} reports means of three runs on the same split indices: $88.0 \pm 0.9$ / $57.1 \pm 4.7$ on P19, with re-runs of GraFITi and Warpformer at $90.5$ / $66.0$ and $89.4$ / $64.9$, against $89.1$ / $56.6$ and $88.7$ / $52.4$ in the five-split compilation; we do not mix its three-run values into the five-split comparison. We searched the literature on 30 November 2026 and again on 3 December 2026, including submissions and their public discussions on OpenReview. There, DBGL's authors report a variant without message passing between the patients of an inference batch, DBGL-PI, at $93.6 \pm 0.3$ / $67.0 \pm 1.1$ on split 1 with five seeds~\citep{anon2026dbgltmlr}, above our split-1 AUROC and below our split-1 AUPRC (Table~\ref{tab:p19}). Three ICLR 2027 submissions report P19 values above ours in settings other than the five-split protocol, and we do not mix them into the comparison: ScaleIMTS~\citep{anon2026scaleimts}, which adapts a frozen pretrained Timer encoder, $93.4 \pm 0.2$ / $68.2 \pm 0.3$ over five seeds, in a setting its public discussion does not state and with spreads below every five-split spread in Table~\ref{tab:p19}; NBSF~\citep{anon2026nbsf}, $93.0$ / $62.4$, whose P19 cohort and feature definitions differ from those of the published results it is compared with, as its authors state; and S$^3$M~\citep{anon2026s3m}, $92.4 \pm 0.5$ / $66.7 \pm 1.9$ on one 70/15/15 split of its own 72-hour extraction, 7.3\% of whose test records are positive.

\paragraph{Information classes.} Several compared methods read information that the benchmark records do not hold: ViRe~\citep{yu2026vire} passes rendered records through a frozen CLIP vision encoder pretrained on image--text data, the sequence-and-image method of \citet{chen2025seqimage} a Swin Transformer pretrained on ImageNet-21K, and ViTST~\citep{li2023vitst} a vision transformer pretrained on large image datasets; ISTS-PLM~\citep{zhang2025istsplm} uses pretrained GPT-2 and BERT layers, KEDGN~\citep{luo2024kedgn} a pretrained language model that encodes variable descriptions, and TRIAGE~\citep{jang2026triage} a fine-tuned four-billion-parameter language model; STaRFormer~\citep{forstenhausler2025starformer} encodes the static covariates of P19 and P12 with a pretrained RoBERTa (PAM has none, so its PAM result uses no pretrained component). The tables mark them with $^{\ast}$. Our models are trained only on each split's training records and use no pretrained weights; we report the comparison with both classes and name the best result trained on the benchmark alone separately.

\paragraph{Compute.} All models train on one CPU thread. The five-split protocols ran on an AWS c7i-class Intel Xeon host and the paired study of Appendix~\ref{app:compl} on a second host. One P19 split takes 10--13 minutes, one P12 split 18--23 minutes and one PAM split 2.9--4.9 hours; the three five-split protocols take 24.3 CPU-thread hours, and the development runs of Appendix~\ref{app:dev} with the paired study of Appendix~\ref{app:compl} another 32.8. The four added seeds of the single-split P19 comparison took another 0.72 CPU-thread hours on a third host. The stopped selection experiment of Appendix~\ref{app:pamsel} and two further P12 development experiments are not included in these totals.

\section{Results}\label{sec:results}

\subsection{P19}

\begin{table}[t]
\centering\small
\caption{P19, five Raindrop splits, test AUROC and AUPRC (\%, mean $\pm$ s.d.\ over splits). References as reported: $^a$ compiled by MTM~\citep{zhong2025mtm}; $^b$ by ViRe~\citep{yu2026vire}; $^c$ in its authors' public review discussion~\citep{anon2026dbgltmlr}; others from their own papers. $^{\ast}$Uses an encoder or language model pretrained on external data (Section~\ref{sec:protocol}). The lower block uses a different protocol (one split, five seeds; Section~\ref{sec:protocol}).}
\label{tab:p19}
\begin{tabular}{@{}lrr@{}}
\toprule
Method & AUROC & AUPRC \\
\midrule
GRU-D~\citep{che2018grud}$^a$ & $83.9 \pm 1.7$ & $46.9 \pm 2.1$ \\
SeFT~\citep{horn2020seft}$^a$ & $81.2 \pm 2.3$ & $41.9 \pm 3.1$ \\
mTAND~\citep{shukla2021mtand}$^a$ & $84.4 \pm 1.3$ & $50.6 \pm 2.0$ \\
IP-Net~\citep{shukla2019ipnet}$^a$ & $84.6 \pm 1.3$ & $38.1 \pm 3.7$ \\
Raindrop~\citep{zhang2022raindrop}$^a$ & $87.0 \pm 2.3$ & $51.8 \pm 5.5$ \\
ContiFormer~\citep{chen2023contiformer}$^a$ & $84.4 \pm 2.1$ & $50.4 \pm 4.3$ \\
CoFormer~\citep{wei2023coformer}$^a$ & $89.2 \pm 1.8$ & $57.3 \pm 3.3$ \\
Warpformer~\citep{zhang2023warpformer}$^a$ & $88.7 \pm 2.0$ & $52.4 \pm 4.5$ \\
ViTST~\citep{li2023vitst}$^{a\ast}$ & $89.2 \pm 2.0$ & $53.1 \pm 3.4$ \\
t-PatchGNN~\citep{zhang2024tpatchgnn}$^a$ & $87.0 \pm 1.4$ & $51.5 \pm 5.2$ \\
GraFITi~\citep{yalavarthi2024grafiti}$^a$ & $89.1 \pm 2.6$ & $56.6 \pm 6.3$ \\
TimeCHEAT~\citep{liu2025timecheat} & $89.5 \pm 1.9$ & $56.1 \pm 4.6$ \\
ISTS-PLM~\citep{zhang2025istsplm}$^{\ast}$ & $89.4 \pm 2.2$ & $56.9 \pm 5.0$ \\
STraTS~\citep{tipirneni2021strats}$^b$ & $89.7 \pm 1.8$ & $57.9 \pm 3.3$ \\
STaRFormer~\citep{forstenhausler2025starformer}$^{\ast}$ & $89.4 \pm 1.3$ & $61.3 \pm 3.4$ \\
SPECTRA~\citep{qin2026spectra} & $90.0 \pm 2.0$ & $59.4 \pm 5.4$ \\
MTM~\citep{zhong2025mtm} & $90.3 \pm 2.0$ & $58.3 \pm 5.3$ \\
Sequence + image SSL~\citep{chen2025seqimage}$^{\ast}$ & $\mathbf{91.6 \pm 0.9}$ & $59.6 \pm 1.3$ \\
ViRe~\citep{yu2026vire}$^{\ast}$ & $91.3 \pm 2.0$ & $62.3 \pm 4.3$ \\
Gradient-boosted trees on raw channel summaries (ours, Section~\ref{sec:compl}) & $91.4 \pm 2.0$ & $61.8 \pm 4.1$ \\
\ours{} (62,681 parameters) & $\mathbf{91.6 \pm 2.2}$ & $\mathbf{63.9 \pm 3.9}$ \\
\midrule
\multicolumn{3}{@{}l}{\emph{One split (split 1), five seeds}} \\
KEDGN, variable names~\citep{luo2024kedgn}$^{\ast}$ & $92.3 \pm 1.0$ & $62.5 \pm 0.7$ \\
Hi-Patch~\citep{luo2025hipatch} & $92.1 \pm 0.4$ & $61.1 \pm 2.1$ \\
DBGL~\citep{chen2026dbgl} & $93.3 \pm 0.5$ & $66.3 \pm 1.4$ \\
DBGL-PI, without message passing between patients$^c$ & $\mathbf{93.6 \pm 0.3}$ & $67.0 \pm 1.1$ \\
\ours{}, split 1 (protocol configuration, five seeds) & $93.4 \pm 0.5$ & $\mathbf{69.7 \pm 1.8}$ \\
\bottomrule
\end{tabular}
\end{table}

Table~\ref{tab:p19} lists published results on the five P19 splits. Our AUPRC is the highest published: $1.6$ points above ViRe, which uses a pretrained CLIP vision encoder, and $4.5$ and $5.6$ points above SPECTRA and MTM, the best results trained on the benchmark alone; at 4.2\% prevalence AUPRC separates the methods more than AUROC does. Our AUROC equals the best published value, which also comes from a model with a pretrained image encoder, and is $1.3$ points above MTM's $90.3$, the best in the table among models trained on the benchmark alone. The split-to-split standard deviation of our AUPRC (3.9 points) is larger than the margin over ViRe, with three of five splits above ViRe's mean, and smaller than the margins over SPECTRA and MTM, with every split above both. Per-split values are in Appendix~\ref{app:splits}.

\subsection{What the temporal state and the decaying value vectors add beyond summary statistics}\label{sec:compl}

\begin{table}[t]
\centering\small
\caption{P19, the same five splits, paired. Every arm is trained on the training part, selected on validation and scored once on test. Reported: the protocol model of Table~\ref{tab:p19}. Rerun: an independent rerun of the same configuration and seed on a second host, which on split 4 selected epoch 7 instead of 6 (Appendix~\ref{app:compl}). Differences are the mean $\pm$ s.d.\ across the five splits, with the number of positive differences in parentheses. Each pair of test sets shares 363--410 records.}
\label{tab:compl}
\begin{tabular}{@{}lrr@{}}
\toprule
Arm & AUROC & AUPRC \\
\midrule
Statistics only (temporal state, value vectors zeroed) & $90.0 \pm 1.8$ & $57.2 \pm 3.7$ \\
Boosted trees on raw summaries & $91.4 \pm 2.0$ & $61.8 \pm 4.1$ \\
Full model, reported & $91.6 \pm 2.2$ & $63.9 \pm 3.9$ \\
Full model, rerun & $91.6 \pm 2.2$ & $64.4 \pm 3.9$ \\
\midrule
Reported $-$ statistics only & $+1.6 \pm 0.7$ (5/5) & $+6.7 \pm 1.4$ (5/5) \\
Reported $-$ trees & $+0.2 \pm 0.7$ (3/5) & $+2.1 \pm 1.8$ (4/5) \\
Rerun $-$ statistics only & $+1.6 \pm 0.7$ (5/5) & $+7.2 \pm 0.5$ (5/5) \\
Rerun $-$ trees & $+0.2 \pm 0.7$ (3/5) & $+2.6 \pm 1.4$ (5/5) \\
\bottomrule
\end{tabular}
\end{table}

\begin{figure}[t]
\centering
\includegraphics[width=\linewidth]{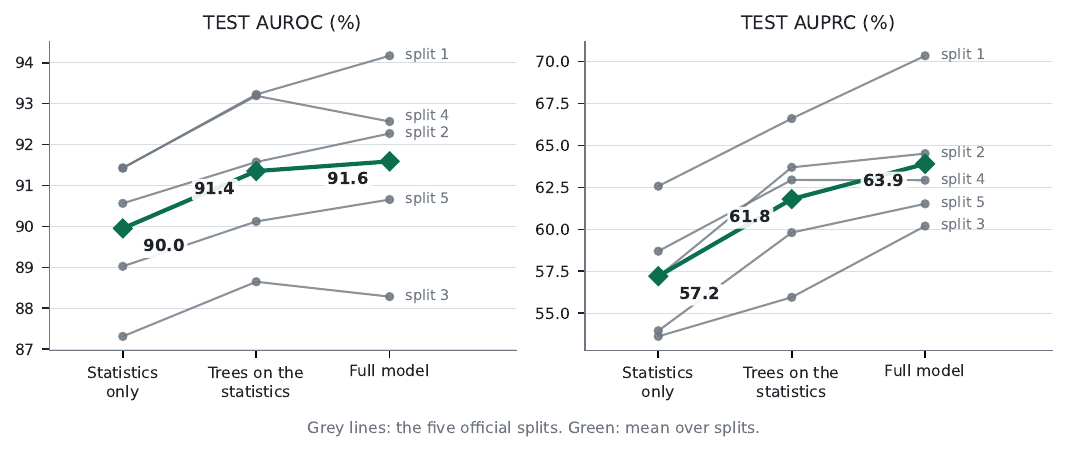}
\caption{P19 test scores of the three arms of Table~\ref{tab:compl} on each of the five splits; the full model is the reported model. The temporal state and the decaying value vectors add to the statistics on every split. The full model scores above the trees in AUPRC on four of the five splits and level with them on split 4 ($62.92$ against $62.94$); the rerun of Table~\ref{tab:compl} scores above them on all five.}
\label{fig:paired}
\end{figure}

Our first design, a temporal model with decaying value slots but without statistics, did no better on P12 validation data than gradient-boosted trees on per-channel summaries (Appendix~\ref{app:dev}). Carrying exact statistics in the slots moved it past them on AUROC ($87.2$ against $86.7$), though not on AUPRC ($57.5$ against $58.5$). This raises the question whether the final model is a statistics reader. Table~\ref{tab:compl} and Figure~\ref{fig:paired} answer it with three arms on the same splits:
\begin{itemize}
    \item Zeroing the temporal-state outputs and value vectors at the readout and retraining costs $1.6$ AUROC and $6.7$ AUPRC points, on every split ($1.6$ and $7.2$ against the rerun).
    \item Gradient-boosted trees~\citep{pedregosa2011sklearn} on raw per-channel summaries, with only the number of boosting rounds chosen on validation, are a strong baseline: $91.4$ / $61.8$; among the five-split results of Table~\ref{tab:p19} only two published deep models exceed them, each on one metric and both with pretrained image encoders. The reported model is level with them on AUROC and ahead on AUPRC by $2.1 \pm 1.8$ points on average, on four of five splits (on split 4, $62.92$ against $62.94$); the rerun, which selected epoch 7 instead of 6 on split 4, is ahead on every split, by $2.6 \pm 1.4$ points. Our neural statistics reader alone scores $90.0$ AUROC against the trees' $91.4$. The within-network ablation isolates the contribution of the temporal state and value vectors.
\end{itemize}

\subsection{PAM and P12}\label{sec:pam}

\begin{table}[t]
\centering\small
\caption{PAM, five splits, test accuracy and macro F1 (\%). $^a$ compiled by MTM; $^b$ by ViRe. $^{\ast}$Uses an encoder pretrained on external data (Section~\ref{sec:protocol}). Precision and recall per split are in Appendix~\ref{app:splits}.}
\label{tab:pam}
\begin{tabular}{@{}lrr@{}}
\toprule
Method & Accuracy & F1 \\
\midrule
Raindrop~\citep{zhang2022raindrop}$^a$ & $88.5 \pm 1.5$ & $89.8 \pm 1.0$ \\
Warpformer~\citep{zhang2023warpformer}$^a$ & $94.2 \pm 2.3$ & $94.9 \pm 2.2$ \\
ViTST~\citep{li2023vitst}$^{a\ast}$ & $95.8 \pm 1.3$ & $96.5 \pm 1.2$ \\
GraFITi~\citep{yalavarthi2024grafiti}$^a$ & $96.0 \pm 0.8$ & $96.1 \pm 0.7$ \\
ISTS-PLM~\citep{zhang2025istsplm}$^{\ast}$ & $96.3 \pm 0.5$ & $96.8 \pm 0.7$ \\
STraTS~\citep{tipirneni2021strats}$^b$ & $96.4 \pm 0.8$ & $95.3 \pm 0.7$ \\
TimeCHEAT~\citep{liu2025timecheat} & $96.5 \pm 0.6$ & $97.0 \pm 0.5$ \\
MTM~\citep{zhong2025mtm} & $97.5 \pm 0.2$ & $97.6 \pm 0.2$ \\
STaRFormer~\citep{forstenhausler2025starformer} & $97.6 \pm 0.9$ & $97.4 \pm 0.3$ \\
SPECTRA~\citep{qin2026spectra} & $97.7 \pm 0.2$ & $98.0 \pm 0.1$ \\
Sequence + image SSL~\citep{chen2025seqimage}$^{\ast}$ & $\mathbf{98.3 \pm 0.3}$ & $\mathbf{98.5 \pm 0.7}$ \\
ViRe~\citep{yu2026vire}$^{\ast}$ & $\mathbf{98.3 \pm 0.6}$ & $98.4 \pm 0.6$ \\
\ours{} (46,316 parameters) & $97.8 \pm 0.7$ & $98.0 \pm 0.8$ \\
\bottomrule
\end{tabular}
\end{table}

\paragraph{PAM.} Without varied training records the model memorised the 4,266 training segments (training loss near zero, validation accuracy 93.4--94.6\%). Random crops and per-channel gains removed that ceiling ($+2.6$ points on split-1 validation data), and the weight average made checkpoint selection stable. The result, $97.8 \pm 0.7$ accuracy and $98.0 \pm 0.8$ F1 (Table~\ref{tab:pam}), is level with SPECTRA ($97.7 \pm 0.2$ / $98.0 \pm 0.1$), the best published five-split result trained on the benchmark alone: our mean accuracy, 97.75\%, exceeds SPECTRA's by less than its printed precision. It is above MTM ($97.5 \pm 0.2$) and every other method in the MTM comparison, and $0.5$ points below the two methods with pretrained image encoders. Four splits score 97.9--98.1\%; split 5 scores 96.4\%, selected at epoch 46 where the other splits were selected at epochs 73--117 (Appendix~\ref{app:pamsel}).

\paragraph{P12.} With the P19 architecture and optimisation settings and a longer epoch budget, P12 in-hospital mortality reaches $87.1 \pm 1.1$ AUROC and $58.5 \pm 2.5$ AUPRC. That is behind MTM ($88.0 \pm 1.0$ / $58.6 \pm 4.1$), the best published five-split result trained on the benchmark alone; behind ViRe ($88.5 \pm 1.2$ / $60.3 \pm 2.4$), TRIAGE~\citep{jang2026triage} ($87.3 \pm 1.2$ / $59.0 \pm 3.4$) and, on AUROC, ISTS-PLM~\citep{zhang2025istsplm} ($87.6 \pm 1.4$ / $57.6 \pm 3.3$), which use pretrained encoders or language models; and ahead of SPECTRA~\citep{qin2026spectra} ($86.3 \pm 0.9$ / $54.4 \pm 3.5$), GraFITi ($86.6$ / $54.8$), Warpformer ($86.5$ / $54.3$) and ViTST ($85.1$ / $51.1$)~\citep{zhong2025mtm}, and of the sequence-and-image method of \citet{chen2025seqimage} ($86.0 \pm 0.3$ / $50.4 \pm 2.1$), which uses a pretrained image encoder. TRIAGE also re-ran baselines on these splits with corrected demographic fields: KEDGN $87.8 \pm 1.1$ (with its pretrained language model), STraTS $87.3 \pm 1.5$, Hi-Patch $87.3 \pm 0.8$ and GRU-D $87.1 \pm 0.9$ AUROC. Under other protocols, GSNF~\citep{gao2026gsnf} reports $87.5 \pm 0.3$ / $56.3$ on five random splits of its own, and DBGL $88.1 \pm 0.4$ / $56.3$ in KEDGN's single-split setting. Validation AUROC on P12 peaks at epochs 2--6; learning rate, weight averaging, crops, further timing inputs and a further learned map of the statistics did not move that peak on split-1 validation data (Appendix~\ref{app:dev}). The model is weakest on sparsely measured stays: by tercile of the number of measurements, split-1 validation AUROC is 0.809, 0.882 and 0.898.

\subsection{Size and cost}

The models have 62,681 (P19), 65,001 (P12) and 46,316 (PAM) parameters. Each update touches the slots of the measured channels only, and the temporal state costs $O(K d)$ per measurement step regardless of the gap. Training runs on one CPU thread in the times given in Section~\ref{sec:protocol}; no GPU was used.

\section{Scope and limitations}\label{sec:limits}

\begin{itemize}
    \item \textbf{Seeds.} Each split has one training run (seed 0); the reported deviation is over splits. For the single-split comparison, seeds 1--4 were added on P19 split 1 with the configuration unchanged, and all five are reported. On split-1 validation data of P12, five seeds of the protocol configuration span 86.5--87.3 AUROC and 56.3--59.1 AUPRC.
    \item \textbf{Margins.} On P19 the AUROC equals the best published value, and the AUPRC margin over the next-best result (1.6 points, ViRe) is smaller than the split-to-split deviation; the margins over SPECTRA and MTM (4.5 and 5.6 points) are larger. Per-split values of the references are not published, so no paired test against them is possible, and the five test sets overlap (each pair shares 363--410 records). On PAM the model is level with SPECTRA, and on P12 it is behind MTM; methods with pretrained image encoders score higher on both.
    \item \textbf{Protocol.} Results under the single-split setting of KEDGN, Hi-Patch and DBGL are not directly comparable to five-split results; our split-1 row in Table~\ref{tab:p19} follows their setting with five seeds on split 1. The PAM release cuts each recording into 600-step windows with 50\% overlap and splits the windows at random~\citep{zhang2022raindrop}, so test windows overlap training windows of the same session; PAM scores here, as for every method on these splits, measure recognition within recorded sessions, not on unseen wearers.
    \item \textbf{Clinical scope.} The labels are those of the public releases, the data are retrospective and from few institutions, and nothing here is a clinical validation. The models are research artifacts, not medical devices.
\end{itemize}

\section{Conclusion}

Keeping an irregular record as measurement events, with per-channel slots that are written only when their channel is measured and a temporal state carried over the actual gaps, gives a 63K-parameter CPU model with the highest published P19 AUPRC under the standard five-split protocol, above models trained on the benchmark alone and models with externally pretrained image encoders. Two observations should travel beyond this model: gradient-boosted trees on per-channel statistics are a stronger P19 baseline than most published deep models and belong in future comparisons, and the temporal state and the decaying value vectors add information those statistics cannot express, as the paired ablation shows on every split.

\section*{AI assistance}
Implementation, experiment execution, result checking and drafting of this paper were carried out with LLM-based coding and writing assistants under the author's direction. The author checked every reported number against the saved result files and takes full responsibility for the content. No LLM is part of the method.

\bibliographystyle{plainnat}

\appendix

\section{Implementation details}\label{app:impl}

\begin{table}[h]
\centering\small
\caption{Configurations of the five-split protocols.}
\label{tab:hparams}
\begin{tabular}{@{}lccc@{}}
\toprule
 & P19 & P12 & PAM \\
\midrule
Width $d$, modes per level $K$, levels & 32, 16, 2 & 32, 16, 2 & 32, 16, 2 \\
Thresholds per channel $J$, value components $d_v$ & 4, 4 & 4, 4 & 4, 4 \\
Statistics per channel & 8 & 8 & 10 \\
Dropout, weight decay & 0.2, $10^{-4}$ & 0.2, $10^{-4}$ & 0.2, $10^{-4}$ \\
Optimiser, learning rate, gradient-norm clip & AdamW, $2\cdot10^{-3}$, 1.0 & same & same \\
Batch size & 128 & 128 & 64 \\
Max.\ epochs, patience & 40, 8 & 60, 12 & 120, 25 \\
Selection metric (validation) & AUROC & AUROC & accuracy \\
Crops, gain s.d., weight average & -- & -- & 80\%, 0.1, 0.999 \\
Parameters & 62,681 & 65,001 & 46,316 \\
\bottomrule
\end{tabular}
\end{table}

Thresholds are initialised evenly in $[-1.5, 1.5]$ with scale $0.5$; angular frequencies from $\mathcal{N}(0, 0.5^2)$; value-vector rates log-spaced between $10^{-2}$ and $10^{0.5}$ per time unit. A never-measured channel has a presence indicator of 0 and default numerical inputs. Time units are hours for P19 and P12 and steps for PAM. Static covariates are standardised with training statistics.

\clearpage
\section{Per-split results}\label{app:splits}

\begin{table}[h]
\centering\small
\caption{Test results per split (\%). P19 and P12: AUROC / AUPRC; PAM: accuracy / F1 and precision / recall.}
\begin{tabular}{@{}lccccc@{}}
\toprule
Split & 1 & 2 & 3 & 4 & 5 \\
\midrule
P19 & 94.2 / 70.3 & 92.3 / 64.5 & 88.3 / 60.2 & 92.6 / 62.9 & 90.7 / 61.5 \\
P12 & 87.5 / 54.8 & 88.5 / 61.4 & 86.4 / 58.5 & 87.7 / 58.0 & 85.6 / 59.6 \\
PAM acc / F1 & 97.9 / 98.3 & 98.1 / 98.3 & 98.1 / 98.4 & 98.1 / 98.5 & 96.4 / 96.6 \\
PAM prec / rec & 98.5 / 98.1 & 98.7 / 97.9 & 98.5 / 98.3 & 98.6 / 98.6 & 96.9 / 96.5 \\
\bottomrule
\end{tabular}
\end{table}

\section{Development record}\label{app:dev}

All development used validation data of split 1, except one P19 variant compared on splits 1--3 (end of this appendix); no test score informed a design choice for the reported configurations.

\begin{table}[h]
\centering\small
\caption{P12, split-1 validation, one seed unless stated. Indented rows modify the closest unindented design above them; the modifications are not cumulative. $^\dagger$Run on a variant that also carries the two PAM statistics (standard deviation and mean absolute change).}
\label{tab:dev}
\begin{tabular}{@{}p{9.2cm}rr@{}}
\toprule
Design & AUROC & AUPRC \\
\midrule
Temporal state + decaying value slots, no statistics (first design, 41,833 parameters) & 86.6 & 54.9 \\
\quad + dropout 0.4, weight decay $10^{-3}$ & 85.9 & 53.1 \\
\quad + learning rate $7\cdot10^{-4}$, weight average 0.995 & 86.0 & 53.6 \\
\textbf{+ statistics in the channel slots (final design, 65,001 parameters)} & \textbf{87.2} & \textbf{57.5} \\
\quad + learning rate $10^{-3}$, weight average 0.995 & 87.1 & 57.4 \\
\quad + learning rate $5\cdot10^{-4}$ & 87.0 & 55.5 \\
\quad + a further learned map of the statistics (145,193 parameters) & 86.8 & 56.4 \\
\quad + further timing inputs per channel and per record$^\dagger$ & 87.2 & 56.9 \\
\quad + random crops 80\% and gains$^\dagger$ & 86.7 & 56.0 \\
Final design, seeds 0--4 & 86.5--87.3 & 56.3--59.1 \\
\midrule
Diagnostic: boosted trees with scikit-learn's default settings (100 rounds, learning rate 0.1) on per-channel summaries and static covariates & 86.7 & 58.5 \\
\bottomrule
\end{tabular}
\end{table}

On P19 split-1 validation data the final design scored 91.9 / 62.6, and adding the further timing inputs (on the variant with the two PAM statistics) gave 91.8 / 62.6. On PAM split-1 validation data: statistics slots alone 94.6\% accuracy; adding standard deviation and mean absolute change 94.0\% (93.4\% at width $d = 48$ with 48 modes, 98,652 parameters); random crops and gains 97.2\% (best epoch 46 of 62, with epoch-to-epoch swings of 94.6--97.2\%); weight average over 120 epochs 98.1\%; a wider model ($d = 48$, 48 modes, 98,652 parameters) 96.8\%; stronger crops (60\%) and gains (0.15) 97.2\%. An additional encoding of the statistics on P19 (validation, splits 1--3) changed AUROC by $+0.3$, $+0.3$, $-0.4$ points and AUPRC by $0.0$, $+0.5$, $-0.9$, and was not adopted.

\section{Statistics diagnostic and paired study}\label{app:compl}

The trees use scikit-learn's histogram gradient boosting~\citep{pedregosa2011sklearn} with learning rate 0.05 on the per-channel count, mean, minimum, maximum, first and last value, last-minus-first change and time of last measurement, plus the static covariates (278 features on P19); the number of rounds is chosen on validation from $\{100, 200, 400, 700, 1000\}$ and the selected model scores test once. The trees take summaries in the original measurement units and the time of the last observation. The neural slots summarize train-normalized, clipped measurements and carry elapsed time since the last observation. The comparison with the neural statistics-only arm holds this encoding fixed. The P12 diagnostic of Table~\ref{tab:dev} uses the same features (297 on P12) with scikit-learn's default settings (100 rounds, learning rate 0.1) and no selection: it is fitted on the split-1 training part and scored on the split-1 validation part. The statistics-only arm is the same network trained from scratch with the temporal-state outputs and value vectors set to zero at the readout, so it reads the statistics, ages, presence indicators and static covariates. The rerun of Table~\ref{tab:compl} retrained the full model on a second host with the same configuration and seed: its AUROC matches the reported model within 0.04 points on every split, and its AUPRC within 0.03 points on four splits. On split 4 its validation selection picked epoch 7 instead of epoch 6; that checkpoint scores 65.5 AUPRC against 62.9, which raises the mean AUPRC from 63.9 to 64.4 and turns the reported model's deficit to the trees on that split ($62.92$ against $62.94$) into a lead. Table~\ref{tab:compl} gives the paired statistics for both models.

\section{PAM checkpoint selection}\label{app:pamsel}

Validation accuracy on PAM moves in steps of 0.19 points (533 records) and saturates near ten errors, so ties are frequent; the protocol keeps the earliest epoch with the best accuracy. After the test results were known we tried selecting on validation cross-entropy instead. On split 2 it selected epoch 45 and scored 97.0\% against 98.1\% for the protocol model; on split 1 it selected epoch 111 instead of 117 and scored the same 97.9\%; a related variant scored 96.6\% on split 5 (protocol model 96.4\%). The experiment was stopped before the remaining splits, and Table~\ref{tab:pam} reports the registered protocol only.

\section{Data and licences}\label{app:data}

The processed P19, P12 and PAM releases of \citet{zhang2022raindrop} are distributed on figshare under CC BY 4.0 (DOIs 10.6084/m9.figshare.19514338, .19514341 and .19514347). The underlying PhysioNet challenge data~\citep{goldberger2000physionet} and PAMAP2~\citep{reiss2012pamap2} are cited above. The processed arrays are pickled NumPy objects; we load them with an allow-list unpickler. Code and configurations: \url{https://github.com/Hypersleep-ai/channel-memories}.

\end{document}